\documentclass[onecolumn]{openhelix}
\usepackage[most]{tcolorbox}

\usepackage{amsmath}
\usepackage{amssymb}
\usepackage{tabularx}
\usepackage{float}
\usepackage{utfsym}
\usepackage{array}
\usepackage{booktabs}
\usepackage{makecell}
\usepackage{xspace}
\usepackage{mathtools}
\usepackage{amsthm}
\usepackage{multirow}
\usepackage{graphicx}

\usepackage{wrapfig}
\usepackage{enumitem}

\usepackage[table]{xcolor}
\definecolor{newyellow}{HTML}{FFD94D}
\definecolor{newgrey}{HTML}{7F7F7F}
\definecolor{newpink}{HTML}{FBCDF4}
\definecolor{realworldoft}{HTML}{029533}
\definecolor{realworldsf}{HTML}{8CC46A}
\definecolor{realworldour}{HTML}{FFC715}
\newcommand{\method}{\texttt{CapVector}}
\usepackage[most]{tcolorbox}
\tcbuselibrary{theorems}
\newtcbtheorem{finding}{Finding}{
  colback=openhelixred!5,
  colframe=openhelixred!60!black,
  fonttitle=\bfseries,
}{find}

\title{CapVector: Learning Transferable Capability Vectors in Parametric Space for Vision-Language-Action Models}

\author[1*]{Wenxuan Song}
\author[2,3*\dagger]{Han Zhao}
\author[1,4]{Fuhao Li}
\author[1]{Ziyang Zhou}
\author[4]{Xi Wang}
\author[5]{Jing Lyu}
\author[2,3]{Pengxiang Ding}
\author[4]{Yan Wang}
\author[3\ddagger]{Donglin Wang}
\author[1\ddagger]{Haoang Li}

\affiliation[1]{HKUST (GZ)}
\affiliation[2]{Zhejiang University}
\affiliation[3]{Westlake University}
\affiliation[4]{Tsinghua University}
\affiliation[5]{Beijing Academy of Artificial Intelligence}

\contribution[*]{Equal Contribution}
\contribution[\dagger]{Project Lead}
\contribution[\ddagger]{Corresponding Author}

\metadata[Code]{\url{https://github.com/OpenHelix-Team/CapVector}}
\metadata[Website]{\url{https://capvector.github.io}}
\metadata[Weights (ready to use)]{\url{https://huggingface.co/haofuly/capvector_models_collection}}

\abstract{

This paper proposes a novel approach to address the challenge that pretrained VLA models often fail to effectively improve performance and reduce adaptation costs during standard supervised finetuning (SFT). 
Some advanced finetuning methods with auxiliary training objectives can improve performance and reduce the number of convergence steps. 
However, they typically incur significant computational overhead due to the additional losses from auxiliary objectives. 
To simultaneously achieve the enhanced capabilities of auxiliary training with the simplicity of standard SFT, we decouple the two objectives of auxiliary-objective SFT within the parameter space, namely, enhancing general capabilities and fitting task-specific action distributions. 
To deliver the goal, we only need to train the model to converge on a small-scale task set using two distinct training strategies, resulting in two finetuned models.
The parameters' difference between the two models can then be interpreted as \textit{capability vectors} provided by auxiliary objectives. These vectors are then merged with pretrained parameters to form a capability-enhanced meta model. Moreover, when standard SFT is augmented with a lightweight orthogonal regularization loss, the merged model attains performance comparable to auxiliary finetuned baselines with reduced computational overhead. Internal and external experiments demonstrate that our capability vectors (1) are effective and versatile across diverse models, (2) can generalize to novel environments and embodiments out of the box.
}

\correspondence{%
\begin{tabular}[t]{@{}l@{\ }l@{}}
Wenxuan Song at & \email{songwenxuan0115@gmail.com}\\
Han Zhao at & \email{zhaohan34@westlake.edu.cn}
\end{tabular}%
}

\begin{document}

\maketitle

\section{Introduction}
\label{sec:introduction}
Vision–Language–Action (VLA) models have become a dominant paradigm in current research on robotic foundation models. They map multimodal perception into executable robotic control, exhibiting a certain degree of language following and visual generalization ability. Similar to Large Language Models (LLMs) \citep{agarwal2025gpt,yang2025qwen3}, training VLAs typically consists of two processes: (1) A pre-training process that allows the model to learn the mapping relation between multimodal input and action output. This process is conducted on large-scale robotic datasets and costs thousands of GPU hours. (2) A finetuning process that allows the model to fit the specific task structure. 


However, recent studies have revealed that pre-trained models do not exhibit the expected strong generalization capability on certain complex downstream tasks. That is, merely collecting a small number of demonstrations and performing standard supervised finetuning (SFT) is often insufficient for the model to quickly adapt to the task and achieve performance significantly superior to training from scratch \citep{kim2025openvla,kim2025fine,black2024pi_0, bjorck2025gr00t}.
Several approaches aim to augment the standard SFT with an auxiliary objective. By designing auxiliary training objectives \citep{flare, song2025reconvla, li2025spatial, laravla, liu2026last} aimed at enhancing specific foundational capabilities, this paradigm enables the model to not only fit the target task's action distribution but also strengthen the corresponding foundational abilities (\textit{e.g.}, spatial perception and multimodal reasoning). With appropriately designed auxiliary objectives, models can significantly reduce the number of training steps required for convergence and achieve downstream performance that surpasses that of standard SFT.

\begin{wrapfigure}{r}{0.45\textwidth}  
    \vspace{-9pt} %
    \centering
    \includegraphics[width=\linewidth]{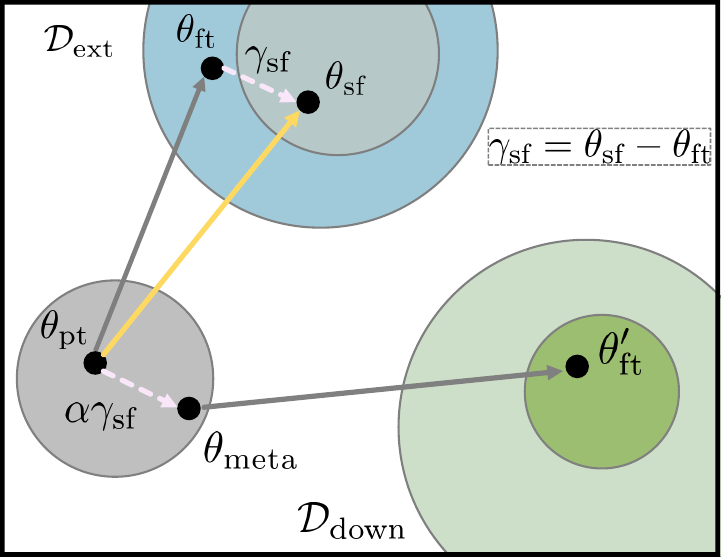} 
    \caption{\textbf{Illustration of our \method.} 
    The \textcolor{newgrey}{\textbf{grey line}} indicates the regular finetuning from pretrained models to finetuned models, the \textcolor{newyellow}{\textbf{yellow line}} indicates the specifically designed finetuning with auxiliary objectives, and the \textcolor{newpink}{\textbf{pink dashed line}} denotes the capability vectors.
    The outer circle denotes the region of the parameter space that yields workable performance on the data manifold, while the inner circle represents the subregion associated with superior performance. We observe that by extracting capability vectors from $\mathcal{D}_{\text{ext}}$ and merging them to obtain a capability-enhanced meta model $\theta_\textnormal{meta}$, the performance on $\mathcal{D}_{\text{down}}$ is improved.
    Variables are defined in \Cref{sec:method}.
    }
    \label{fig:teaser} 
\end{wrapfigure}



Despite the above strengths, these approaches have obvious drawbacks: auxiliary objectives often introduce extra modules and additional forward passes. For example, the 3D Foundation Model in Spatial Forcing \citep{li2025spatial} requires additional computation to obtain aligned targets during training, and LaRA-VLA \citep{laravla} requires training of the latent chain-of-thought tokens, which incurs extra computational overhead. As the number of downstream tasks and the scale of data grow, this overhead gradually becomes prohibitive (Appendix~\Cref{appendix:sf_overhead}).
This naturally motivates the following question: \textbf{Can the beneficial properties $s$, induced by carefully designed finetuning procedures, be transferred into the pretrained model itself, such that the model inherently possesses $s$?} If so, one could rely solely on standard SFT to inherit the same training efficiency and performance improvements, without incurring additional overhead.

The answer to this question is \textbf{yes}. Drawing inspiration from the concept of task vectors \citep{ilharco2022editing}, we posit the following assumption: \textit{The two gains obtained during the training process---namely, the improvement of general capabilities and the enhancement of task-specific action fitting accuracy---can be decoupled. Furthermore, the changes in the model after training can be seen as a linear combination of parameter vectors that reflect these two characteristics.} 
Based on the assumption, we can acquire two sets of finetuned model parameters by applying the auxiliary-objective SFT and the standard SFT method to the same downstream task, respectively. The difference between these two sets of parameters can be interpreted as the \textit{capability vectors} (\method). These can then be integrated into the pretrained backbone through arithmetic operations, thereby achieving model merging. 
The whole process is shown in \Cref{fig:teaser}.
While prior work in this field has primarily focused on obtaining an off-the-shelf specialist model via merging \citep{chenbring, fu2025mergevla, yadav2025robust}, it remains unclear whether such techniques can be employed to produce a better generalist model that is more suitable for arbitrary downstream finetuning while also delivering superior performance.
After capability extraction, a lightweight orthogonal regularization loss is needed during downstream finetuning to prevent forgetting of the capability vectors. The detailed implementation is described in~\Cref{sec:method}.


In experiments, we focus on investigating the extractable capabilities and the underlying training mechanism of this approach. 
Extensive experiments demonstrate that the merged meta model can achieve performance and training efficiency comparable to SFT methods with auxiliary objectives across multiple downstream tasks. 
Furthermore, we validate the versatility of \method~through the experiments on diverse VLA architectures and SFT strategies.
After validating the effectiveness, we derive empirical conclusions from a series of experiments on what types of downstream tasks are suitable for extracting high-quality capability vectors. 
Finally, internal and external experiments in the real world demonstrate its practicality and generalization to novel environments and embodiments out of the box.

In summary, our core contributions are as follows:

\begin{itemize}
    \item We define and introduce the concept of the capability vector, which represents the gain in general capabilities acquired during finetuning with auxiliary objectives in the form of model parameters. By merging these capability vectors with the pretrained model, we obtain a capability-enhanced meta model.
    \item Based on the meta model, we only need to make minimal modifications to standard SFT by introducing an orthogonal regularization loss to mitigate forgetting. This achieves both the simplicity of standard SFT and the high performance of auxiliary-objective finetuning during downstream training.
    \item Extensive experiments demonstrate our \method's effectiveness and efficiency as a general learning strategy on various tasks, environments, and models.
    
\end{itemize}
\section{Capability Vectors~(\method)}
\label{sec:method}

Our method consists of two stages.
Before training, we transfer the capability vectors derived from the auxiliary-objective SFT, thereby obtaining an enhanced meta model that inherits the desired properties.
During training, to adapt the model to downstream tasks without degrading these properties, we introduce a regularization strategy in orthogonal subspaces.

\subsection{Problem Formulation}
\label{sec:prob_form}
Assume we have a pretrained VLA model $\theta_\text{pt} \in \mathbb{R}^d$ and a multi-task extensive dataset $\mathcal{D}_\text{ext}$, whose included tasks are referred to as capability extraction tasks. These tasks are specifically designed not for downstream performance, but to induce and expose particular model capabilities through parameter variations during finetuning. We denote the extracted \textit{capability vectors} as $\gamma_\text{ao} \in \mathbb{R}^d$. Our goal is to obtain a more capable meta model $\theta_{\text{meta}} = \theta_{\text{pt}} + \gamma_\text{ao}$ that is superior to the pretrained model by acquiring general capability vectors on a small-scale set of capability extraction tasks. That is, given a downstream task dataset $\mathcal{D}_{\text{down}}$ for evaluation, under consistent training settings, the model obtained by finetuning $\theta_{\text{meta}}$ on $\mathcal{D}_{\text{down}}$ achieves better performance than the model obtained by finetuning $\theta_{\text{pt}}$.

\subsection{Before Training: Capability Vectors Transferring}
First, we consider employing standard SFT on the data in $\mathcal{D}_\text{ext}$, resulting in the finetuned model $\theta_\textnormal{ft}$:
\begin{equation}
\theta_\textnormal{ft} = \theta_\textnormal{pt} + \Delta_\textnormal{ft}.
\label{eq:1}
\end{equation}
We denote $\Delta_\textnormal{ft}$ as the parameter difference between the pretrained and the finetuned model. 

Next, we consider the scenario of extracting capability vectors from SFT methods with auxiliary objectives, such as Spatial Forcing \citep{li2025spatial} that aligns intermediate visual embeddings of VLAs with geometric representations produced by pretrained 3D foundation models to enhance spatial perception, and LaRA-VLA \citep{laravla} that internalises multimodal chain-of-thought into continuous latent representations to enhance long-horizon reasoning capabilities.

We denote the model $\theta_\textnormal{ao}$ finetuned by these auxiliary-objective SFT methods as 
\begin{equation}
    \theta_\textnormal{ao} = \theta_\textnormal{pt} + \Delta_\textnormal{ao} = \theta_\textnormal{pt} + \delta_\textnormal{ao} + \gamma_\textnormal{ao},
\label{eq:2}
\end{equation}
where $\delta_\textnormal{ao}$ denotes the vectors for task-specific action learning, and $\gamma_\textnormal{ao}$ denotes the capability vectors obtained from the auxiliary objective. 
When the finetuning setting is consistent between $\theta_\textnormal{ft}$ and $\theta_\textnormal{ao}$, we assume that the task-relevant vectors can be approximately considered the same, \textit{i.e.}, $\Delta_\textnormal{ft} = \delta_\textnormal{ao}$.
This assumption is empirically supported by the massive experiments below.
Thus, given \Cref{eq:1} and \Cref{eq:2}, we can extract the individual $\gamma_\textnormal{ao}$ by
\begin{equation}
    \gamma_\textnormal{ao} = (\theta_\textnormal{pt} + \delta_\textnormal{ao} + \gamma_\textnormal{ao}) - (\theta_\textnormal{pt} + \Delta_\textnormal{ft}) = \theta_\textnormal{ao} - \theta_\textnormal{ft}.
\label{eq:3}    
\end{equation}
This indicates that we can extract the capability vectors by simply conducting parameter arithmetic between two models finetuned with different strategies.
Then, to achieve our goal of transferring the properties of $\theta_\textnormal{ao}$ to $\theta_\textnormal{pt}$, we merge the capability vectors $\gamma_\textnormal{ao}$ and $\theta_\textnormal{pt}$ and get the capability-enhanced meta model with properties:
\begin{equation}
    \theta_\textnormal{meta} = \theta_\textnormal{pt} +    \alpha\gamma_\textnormal{ao},
\label{eq:cap_merge}    
\end{equation}
where $\alpha$ denotes vector weights. This provides a better initialization for further performing finetuning on any new tasks: 
\begin{equation}
\theta'_\textnormal{ft}=\theta_\textnormal{meta}+\Delta'_\textnormal{ft}.
\label{eq:4}
\end{equation}

\subsection{During Training: Regularization in Orthogonal Subspaces}

While we have transferred the properties to the pretrained model, there is an obvious question: \textit{how to retain the properties during regular finetuning?}

Because the capability vectors and the obtained meta model share the same parametric space, the parameters of the meta model undergo updates within the shared parametric space. Without the auxiliary supervision, the standard SFT can harm the properties, and this phenomenon can be more harmful with more training steps.


Some previous work \citep{o-lora} utilizes orthogonal regularization to maintain the model's performance and continue to learn new tasks. In our case, we aim to keep the orthogonality between the capability vectors $\gamma_\textnormal{ao}$ and $\Delta'_\textnormal{ft}$ to prevent interference. Our fundamental insight is rooted in the nature of finetuning: the parameter changes are not mere numerical adjustments but encapsulate crucial model update directions. Thus, orthogonality needs to satisfy:
\begin{equation}
\langle {\gamma_\textnormal{ao}}^{(p)},{\Delta'_\textnormal{ft}}^{(p)} \rangle = 0, \forall {\gamma_\textnormal{ao}}^{(p)} \in \mathcal{\gamma_\textnormal{ao}}, {\Delta'_\textnormal{ft}}^{(p)} \in \\{\Delta'_\textnormal{ft}},
\end{equation}
where $p$ denotes a parameter in the capability vectors and task vectors.
Therefore, our orthogonal regularization loss is defined as:
\begin{equation}
\label{eq:orth_loss}
\mathcal{L}_\textnormal{orth}(\gamma_\textnormal{ao}, \Delta'_\textnormal{ft}) =\sum_p\sum_{i,j} \lvert
{\gamma_\textnormal{ao}}^{(p)}_{ij}{\Delta'_\textnormal{ft}}^{(p)}_{ij}\rvert
\end{equation}
where $i,j$ denote the element at the $i$-th row and $j$-th column of the matrix.
The total training loss is:
\begin{equation}
\label{eq:reg_loss_weight}
\mathcal{L}=\mathcal{L}_\textnormal{action} + \lambda \mathcal{L}_\textnormal{orth},
\end{equation}
where $\lambda$ is the weight of the orthogonality loss. 
Please note that the extra overhead induced by orthogonality loss is slight, as quantized in Appendix~\Cref{appendix:orth_loss}.
For Low-Rank Adaptation (LoRA) tuning, we only calculate $\mathcal{L}_\textnormal{orth}$ between the matrix $A$ in LoRA.
This is because they represent the updating direction of the model, and matrix $B$ serves as linear weighting coefficients for matrix $A$ \citep{buyukakyuz2024olora}.

\section{Experiments}
\label{sec:experiments}

In this section, we evaluate the effectiveness of our \method~and offer several findings by investigating the following research questions (RQs):
\begin{itemize}[leftmargin=2em]
    \item \textbf{RQ1:} Can \method~effectively transfer capabilities in the domain? How does the design of the loss function and the choice of hyperparameters contribute to the performance?
    \textit{\textbf{(In-distribution Effectiveness)}}
    \item \textbf{RQ2:} Are the extracted capability vectors task-irrelevant? Do they exhibit out-of-domain transferability? \textit{\textbf{(Out-of-distribution Effectiveness \& Generalization)}}
    \item \textbf{RQ3:} Is \method~consistently effective and efficient on various VLA architectures?
    Can it transfer diverse capabilities (\textit{e.g.}, spatial perception and multimodal reasoning) of different auxiliary-objectives SFT? \textbf{\textit{(Versatility)}}
    \item \textbf{RQ4:} What is the determinant to obtain the capability vectors with high qualities? \textit{\textbf{(Mechanism)}}
    \item \textbf{RQ5:} Can \method~realize sim-to-real transfer, \textit{i.e.}, are the capability vectors obtained from simulated environments still effective in the real world? 
    Can \method~work across robot embodiments and real-world scenes out of the box?
    \textit{\textbf{(Real-world Performance \& Practicality)}}

\end{itemize}

\begin{table*}[t]
\setlength{\tabcolsep}{5pt}
\footnotesize
\centering
\caption{\textbf{In-distribution Comparison in LIBERO under various training iterations based on Spatial Forcing.} $\theta_\textnormal{ft}$: OpenVLA-OFT. $\theta_\textnormal{ao}$: Spatial Forcing. $\mathcal{D}_\textnormal{ext}$: \{LIBERO-Spatial\}. $\mathcal{D}_\textnormal{down}$: \{LIBERO-Spatial, Object, Goal, Long\}.}
\renewcommand{\tabularxcolumn}[1]{m{#1}}
\begin{tabularx}{\textwidth}{>{\raggedright\arraybackslash}m{0.14\textwidth}>{\raggedright\arraybackslash}X*{4}{>{\centering\arraybackslash}c}>{\columncolor{gray!15}\centering\arraybackslash}c}
\toprule
\textbf{Progress} & \textbf{Method} & \textbf{Spatial} & \textbf{Object} & \textbf{Goal} & \textbf{Long} & \cellcolor{white}\textbf{Average} \\
\midrule
\multirow{4}{*}{5k Steps}
& Spatial Forcing ($\theta_\textnormal{ao}$)              & 93.8\% & 94.8\% & 94.6\% & 66.6\% & 87.5\% \\
& OpenVLA-OFT ($\theta_\textnormal{ft}$)                  & 87.0\% & \textbf{99.8\%} & 92.8\% & 48.8\% & 82.1\% \\
& \method~w/o orthogonal loss   &   \underline{96.0\%}   &   99.0\%   & \textbf{97.4\%} & \underline{68.0\%} &   \underline{90.1\%}   \\
& \method~(ours) & \textbf{98.0\%} & \underline{99.2\%} & \underline{96.6\%} & \textbf{73.0\%} &  \textbf{91.7\%} \\
\midrule
\multirow{4}{*}{1 Epoch}
& Spatial Forcing ($\theta_\textnormal{ao}$)              & \underline{98.4\%} & 99.6\% &   \underline{97.8\%}    & 84.8\% & 95.2\%   \\
& OpenVLA-OFT ($\theta_\textnormal{ft}$)                  & 97.0\% & \underline{99.8\%} & 96.4\% & 70.4\% & 90.9\% \\
& \method~w/o orthogonal loss   &   \underline{98.4\%}   &   \textbf{100.0\%}   &   \textbf{98.0\%}   & \underline{86.0\%} &   \underline{95.6\%}  \\
& \method~(ours) & \textbf{98.6\%} & \underline{99.8\%} & 97.6\% & \textbf{90.0\%} & \textbf{96.5\%} \\
\midrule
\multirow{4}{*}{8 Epochs}
& Spatial Forcing ($\theta_\textnormal{ao}$)              & 93.0\% & \textbf{98.2\%} &   \textbf{98.4\%}   & 87.2\% &   94.2\%   \\
& OpenVLA-OFT ($\theta_\textnormal{ft}$)                  & 92.8\% & \textbf{98.2\%} & \underline{97.8\%} & 87.0\% & 93.9\% \\
& \method~w/o orthogonal loss   &   \underline{97.6\%}   &   97.8\%   &   96.6\%   & \underline{92.2\%} &   \underline{96.1\%}  \\
& \method~(ours) & \textbf{98.0\%} & \underline{98.0\%} & 96.8\% & \textbf{93.6\%} & \textbf{96.6\%} \\
\midrule
\multirow{4}{*}{150k Steps}
& Spatial Forcing ($\theta_\textnormal{ao}$)              & 97.2\% &  \textbf{99.2\%} & \underline{96.8\%} & \underline{94.2\%} & \underline{96.9\%} \\
& OpenVLA-OFT ($\theta_\textnormal{ft}$)                  & 96.8\% & 94.8\% & 92.8\% & 86.2\% & 92.7\% \\
& \method~w/o orthogonal loss   &   \underline{97.4\%}   &   \underline{99.0\%}  & \textbf{97.2\%} & 91.2\% &   96.2\%  \\
& \method~(ours) & \textbf{98.4\%} &  98.4\% & \underline{96.8\%} &  \textbf{94.8\%} & \textbf{97.1\%} \\
\bottomrule
\end{tabularx}
\label{tab:libero}
\vspace{-0.3cm}
\end{table*}


\subsection{Experimental Settings}
\textbf{Simulated Environments.}
We evaluate our method on two representative simulated benchmarks, LIBERO \citep{liu2023libero} and RoboTwin 2.0 \citep{chen2025robotwin}. 
LIBERO is a widely used benchmark built on Robosuite \citep{zhu2020robosuite}.
It consists of four suites (Spatial, Object, Goal, Long), each comprising 10 tasks. Success rates are reported with 500 rollouts per suite across 3 random seeds.
RoboTwin 2.0 \citep{chen2025robotwin} is a bimanual manipulation benchmark built on Sapien \citep{xiang2020sapien}. 
In this paper, we focus on 10 tasks with clean backgrounds as target datasets and run 100 rollouts per task to calculate success rates.
We also utilize another 5 tasks with clean backgrounds and randomized backgrounds individually as capability extraction tasks in \Cref{sec:determinants}.

\textbf{Base Models.}
We choose three representative VLAs, OpenVLA-OFT \citep{kim2025fine}, StarVLA \citep{starvla}, and $\pi_{0.5}$ \citep{black2025pi} as our regular SFT backbones. We choose two auxiliary-objective SFT methods, Spatial Forcing \citep{li2025spatial} and LaRA-VLA \citep{laravla}, as introduced in~\Cref{appendix:baseline}.
Following official settings, we use LoRA tuning for OpenVLA-OFT and full tuning for StarVLA and $\pi_{0.5}$. 

\textbf{Training Details.} 
All experiments are conducted on NVIDIA H100 GPUs, with 1 GPU used for OpenVLA-OFT, 8 GPUs for StarVLA, and 4 GPUs for $\pi_{0.5}$. Per-device batch size is set to 8 for OpenVLA-OFT, 16 for StarVLA, and 32 for $\pi_{0.5}$. 
Training step is set to 150k for OpenVLA-OFT, 20k for StarVLA, and 60k for $\pi_{0.5}$.
As shown in Section~\ref{sec:prob_form}, we denote the training set of $\theta_\textnormal{ao}$ and $\theta_\textnormal{ft}$ as $\mathcal{D}_\text{ext}$, and denote the training set of $\theta'_\textnormal{ft}$ as $\mathcal{D}_\text{down}$.

\subsection{In-distribution (ID) Study (RQ1)}
\label{sec:id}

\noindent
\textbf{Settings.} 
The following settings are considered for comparison: $\mathcal{D}_\textnormal{ext} = $\{LIBERO-Spatial\} and $\mathcal{D}_\textnormal{ext} = $\{LIBERO-Spatial, Object, Goal, Long\}. We compare our \method~with $\theta_\textnormal{ft}$~(OpenVLA-OFT) and $\theta_\textnormal{ao}$~(Spatial Forcing). Please note that \method~is trained from $\theta_{\textnormal{meta}}$ through standard SFT, identical to that applied to $\theta_{\textnormal{ft}}$.

\noindent
\textbf{\textit{Finding 1:}} \method~\textbf{\textit{inherits the efficiency and effectiveness from $\theta_\textnormal{ao}$.}}
For ID transferring, \Cref{tab:libero} shows that our \method~yields comparative or even higher success rates over Spatial Forcing on all training steps and all tasks. This indicates that the capability vectors are implicit representations of the extra spatial capabilities of Spatial Forcing, and simply merging parameters successfully transfers these capabilities.
Additionally, with only 5k training steps, our \method~achieves a substantially higher success rate than OpenVLA-OFT, despite both models being trained with regular finetuning, indicating that \method~inherits the training efficiency of Spatial Forcing.

\begin{figure}[t]  
    \centering
    \includegraphics[width=1\linewidth]{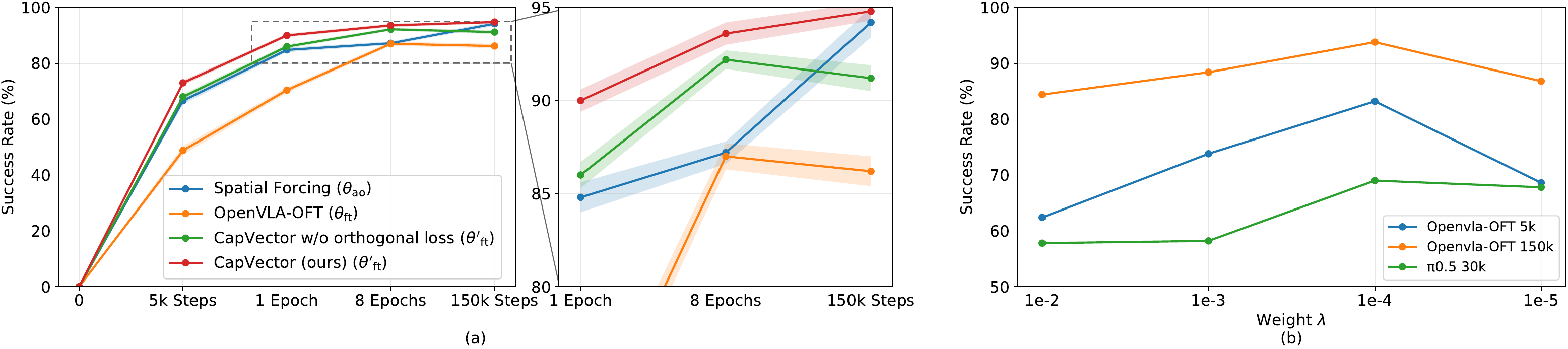} 
    \vspace{-0.6cm}
      \caption{\textbf{(a) Success rates vs. training iterations} on LIBERO-Long for Spatial Forcing ($\theta_\textnormal{ao}$), OpenVLA-OFT ($\theta_\textnormal{ft}$), and \method~($\theta'_\textnormal{ft}$). The shadow
      represents the standard deviation across 3 seeds.
      \textbf{(b) Ablation of the orthogonal regularization weight $\lambda$.} 
      }
    \label{fig:mix_fig_libero_openvla} 
\end{figure}

\noindent
\textbf{\textit{Finding 2: The orthogonal loss is critical for maintaining the capability of the capability vectors.}}
As shown in \Cref{fig:mix_fig_libero_openvla}, while the performances of \method~w/o orthogonal loss are consistently over Spatial Forcing on 5k steps, 1 epoch, and 8 epochs, it can not match Spatial Forcing on the 150k steps, which represents abundant training steps. This indicates that the pre-injected capabilities are updated and reduced during the regular finetuning process, finally resulting in capability degradation.

When the orthogonal loss is incorporated to retain the injected capabilities and constrain the model updating on the new direction, the capability degradation is largely mitigated.
\Cref{tab:libero} shows that \method~with orthogonal loss has a clear performance improvement over the baseline without it, and is still superior to Spatial Forcing with 150k training steps.

\noindent
\textbf{Ablation of Hyperparameters.}
$\lambda$ is used to control the weight of the orthogonal loss in \cref{eq:reg_loss_weight}.
As shown in \cref{fig:mix_fig_libero_openvla}, the model performs best under $\lambda$~=~1e-4, which is the default setting in all other experiments.
Ablation of vector weights $\alpha$ is shown in \Cref{appendix:ablation}.

\begin{table*}[t]
    \centering
    \caption{\fontsize{9}{11}\selectfont\textbf{Comparison in RoboTwin 2.0 benchmark.}
    $\theta_\textnormal{ft}$: \{OpenVLA-OFT\}. $\theta_\textnormal{ao}$: \{Spatial Forcing\}.
    $\mathcal{D}_\textnormal{ext}$: \{LIBERO-Spatial\}.
    $\mathcal{D}_\textnormal{down}$: \{RoboTwin 2.0 10 tasks with clean settings.\}}
    \label{tab:robotwin}

    \fontsize{6.7}{8}\selectfont
    \renewcommand{\arraystretch}{1.05}
    \renewcommand{\tabularxcolumn}[1]{m{#1}}
    
    \setlength{\tabcolsep}{3.5pt}
    \begin{tabularx}{\linewidth}{>{\raggedright\arraybackslash}m{0.26\linewidth}*{11}{>{\centering\arraybackslash}X}}
    \toprule
    \shortstack[c]{\fontsize{8}{8}\selectfont\bfseries Method} &
    \shortstack[c]{\fontsize{4.8}{5.8}\selectfont\bfseries Turn\\[-0.8pt]\fontsize{4.8}{5.8}\selectfont\bfseries switch} &
    \shortstack[c]{\fontsize{4.8}{5.8}\selectfont\bfseries Hand-\\\fontsize{4.8}{5.8}\selectfont\bfseries over\\[-0.8pt]\fontsize{4.8}{5.8}\selectfont\bfseries block} &
    \shortstack[c]{\fontsize{4.8}{5.8}\selectfont\bfseries Hand-\\\fontsize{4.8}{5.8}\selectfont\bfseries over\\[-0.8pt]\fontsize{4.8}{5.8}\selectfont\bfseries mic} &
    \shortstack[c]{\fontsize{4.8}{5.8}\selectfont\bfseries Place\\[-0.8pt]\fontsize{4.8}{5.8}\selectfont\bfseries shoe} &
    \shortstack[c]{\fontsize{4.8}{5.8}\selectfont\bfseries Pick\\[-0.8pt]\fontsize{4.8}{5.8}\selectfont\bfseries dual\\[-0.8pt]\fontsize{4.8}{5.8}\selectfont\bfseries bottles} &
    \shortstack[c]{\fontsize{4.8}{5.8}\selectfont\bfseries Place\\[-0.8pt]\fontsize{4.8}{5.8}\selectfont\bfseries object\\[-0.8pt]\fontsize{4.8}{5.8}\selectfont\bfseries basket} &
    \shortstack[c]{\fontsize{4.8}{5.8}\selectfont\bfseries Put\\[-0.8pt]\fontsize{4.8}{5.8}\selectfont\bfseries bottles\\[-0.8pt]\fontsize{4.8}{5.8}\selectfont\bfseries dustbins} &
    \shortstack[c]{\fontsize{4.8}{5.8}\selectfont\bfseries Place\\[-0.8pt]\fontsize{4.8}{5.8}\selectfont\bfseries phone\\[-0.8pt]\fontsize{4.8}{5.8}\selectfont\bfseries stand} &
    \shortstack[c]{\fontsize{4.8}{5.8}\selectfont\bfseries Put\\[-0.8pt]\fontsize{4.8}{5.8}\selectfont\bfseries object\\[-0.8pt]\fontsize{4.8}{5.8}\selectfont\bfseries cabinet} &
    \shortstack[c]{\fontsize{4.8}{5.8}\selectfont\bfseries Stack\\[-0.8pt]\fontsize{4.8}{5.8}\selectfont\bfseries bowls\\[-0.8pt]\fontsize{4.8}{5.8}\selectfont\bfseries two} &
    \shortstack[c]{\fontsize{8}{5.8}\selectfont\bfseries Avg.} \\
    \midrule
    {\fontsize{5.8}{8.2}\selectfont OpenVLA-OFT} &
    33.0\% & 1.0\% & 4.0\% & 2.0\% & 1.0\% & 7.0\% & 1.0\% & 1.0\% & 9.0\% & 8.0\% & 6.7\% \\
    {\fontsize{5.8}{8.2}\selectfont + Spatial Forcing} &
    \textbf{47.0\%} & \textbf{23.0\%} & \textbf{100.0\%} & 2.0\% & 17.0\% & 24.0\% & 15.0\% & \textbf{17.0\%} & \textbf{23.0\%} & \textbf{63.0\%} & \textbf{33.1\%} \\
    {\fontsize{5.8}{8.2}\selectfont \method~($\mathcal{D}_\textnormal{ext}$ = Spatial) (ours)} &
    33.0\% & 1.0\% & 99.0\% & \textbf{13.0\%} & \textbf{23.0\%} & \textbf{39.0\%} & \textbf{22.0\%} & 12.0\% & 17.0\% & 59.0\% & 31.8\% \\
    {\fontsize{5.8}{8.2}\selectfont \method~($\mathcal{D}_\textnormal{ext}$ = Long)} &
    18.0\% & 12.0\% & 99.0\% & 13.0\% & 6.0\% & 33.0\% & 4.0\% & 5.0\% & 21.0\% & 26.0\% & 23.7\% \\
    {\fontsize{5.8}{8.2}\selectfont \method~($\mathcal{D}_\textnormal{ext}$ = 90)} &
    36.0\% & 1.0\% & 0.0\% & 2.0\% & 1.0\% & 11.0\% & 3.0\% & 0.0\% & 9.0\% & 7.0\% & 9.4\% \\
    \midrule
    \rowcolor{gray!15}
    {\fontsize{5.8}{8.2}\selectfont $\pi_{0.5}$} &
    3.0\% & \textbf{44.0\%} & 9.0\% & 3.0\% & 27.0\% & 7.0\% & 12.0\% & 1.0\% & 34.0\% & 11.0\% & 15.1\% \\
    \rowcolor{gray!15}
    {\fontsize{5.8}{8.2}\selectfont + Spatial Forcing} &
    \textbf{5.0\%} & 0.0\% & \textbf{15.0\%} & 19.0\% & 29.0\% & \textbf{18.0\%} & \textbf{32.0\%} & 27.0\% & 72.0\% & 19.0\% & \textbf{23.6\%} \\
    \rowcolor{gray!15}
    {\fontsize{5.8}{8.2}\selectfont \method~(Merged VLM)} &
    4.0\% & 1.0\% & 14.0\% & 19.0\% & 27.0\% & \textbf{18.0\%} & \textbf{32.0\%} & 19.0\% & \textbf{73.0\%} & \textbf{22.0\%} & 22.9\% \\
    \rowcolor{gray!15}
    {\fontsize{5.8}{8.2}\selectfont \method~(VLM + Expert) (ours)} &
    \textbf{5.0\%} & 0.0\% & 12.0\% & \textbf{20.0\%} & \textbf{32.0\%} & 15.0\% & 31.0\% & \textbf{34.0\%} & 64.0\% & \textbf{22.0\%} & 23.5\% \\
    \bottomrule
    \end{tabularx}
\end{table*}

\subsection{Out-of-distribution (OOD) Study (RQ2)}
\label{sec:ood}

\begin{figure}[t]
    \centering
    \includegraphics[width=0.6\linewidth]{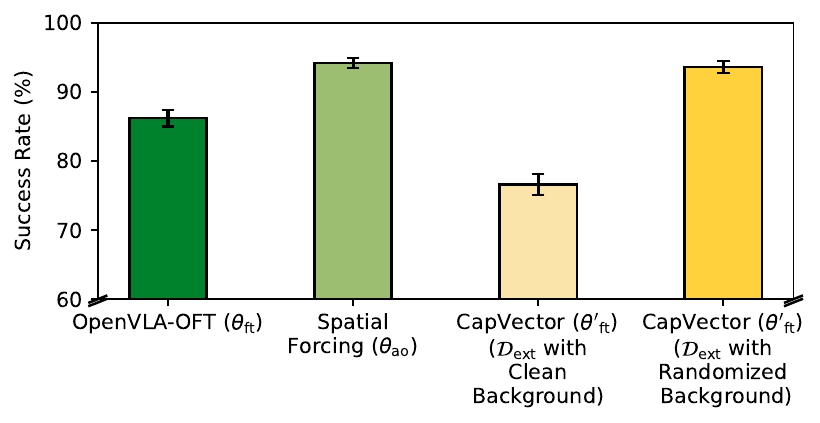}
    \caption{\textbf{Influence of the visual richness of $\mathcal{D}_\textnormal{ext}$.}
    $\mathcal{D}_\textnormal{ext}$: \{RoboTwin 2.0 5 capability extraction tasks\}.
    $\mathcal{D}_\textnormal{down}$: \{LIBERO-Long\}.
    We report the success rates on LIBERO-Long and consider the capability vectors from data with different visual richness.
    The error bars represent the standard error across 3 seeds.}
    \label{fig:richness_compare}
\end{figure}

\noindent
\textbf{\textit{Finding 3: The capability vectors can be seen as task-irrelevant, thus \method~exhibits out-of-distribution transfer
ability.}}
To evaluate the transferring feasibility across domains, we conduct experiments with $\mathcal{D}_\textnormal{ext}$ and $\mathcal{D}_\textnormal{down}$ in different simulated environments.
\Cref{tab:robotwin} shows that for different architectures, capability extraction datasets, and merging strategies, our \method~realizes capability transferring to unseen distribution. Specifically, with the capability vectors $\gamma_{\textnormal{ao}}$ obtained from LIBERO, our \method~always outperforms base models on most tasks of RoboTwin by a clear margin, especially improving success rates from 6.7\% to 31.8\% with OpenVLA-OFT as $\theta_\textnormal{ft}$.
Moreover, it achieves performance comparable to Spatial Forcing.
Furthermore, \Cref{fig:richness_compare} also validates the OOD transferring in the setting that $\mathcal{D}_\textnormal{ext}$ is RoboTwin 2.0 and $\mathcal{D}_\textnormal{down}$ is LIBERO-Long.
The observed improvements in cross-domain success rates provided by capability vectors demonstrate their task-agnostic nature and capacity to facilitate generalized model performance enhancement.

\subsection{Versatility Study (RQ3)}

\noindent
\textbf{\textit{Finding 4:}} \method~\textbf{\textit{demonstrates versatility across pretrained models $\theta_\textnormal{pt}$ with different architectures and diverse auxiliary-objective SFT methods $\theta_\textnormal{ao}$.}}
While the previous experiments have demonstrated the effectiveness of \method~with the OpenVLA-OFT as $\theta_\textnormal{ft}$ and Spatial Forcing as $\theta_\textnormal{ao}$, we further consider other $\theta_\textnormal{ft}$ and $\theta_\textnormal{ao}$. 
Given LIBERO-Spatial as $\mathcal{D}_\textnormal{ext}$, we validate the versatility of \method~on two settings: (1) $\theta_\textnormal{ft}$: \{StarVLA\}, $\theta_\textnormal{ao}$: \{LaRA-VLA\}, and $\mathcal{D}_\textnormal{down}$: \{LIBERO\}. (2) $\theta_\textnormal{ft}$: \{$\pi_{0.5}$\}, $\theta_\textnormal{ao}$: \{Spatial Forcing\}, and $\mathcal{D}_\textnormal{down}$: \{RoboTwin\}.

\Cref{tab:libero,tab:libero_laravla} shows that \method~is effective across distinct auxiliary-objective methods, validating its capacity to extract and transfer diverse foundational capabilities. Specifically, while \cref{tab:libero} highlights its success in extracting geometric comprehensions from Spatial Forcing, \cref{tab:libero_laravla} proves that it can just as effectively capture the multimodal chain-of-thought reasoning abilities internalized by LaRA-VLA. When applied to the StarVLA backbone, \method~achieves an impressive average success rate on LIBERO, outperforming the standard StarVLA baseline and performing comparably to the full LaRA-VLA method. These results indicate that \method~can effectively transfer various capabilities while avoiding the extra computational costs associated with auxiliary SFT methods.

\begin{table}[t]
\centering
\small
\caption{\textbf{LIBERO in-distribution comparison based on LaRA-VLA.}}
\label{tab:libero_laravla}
\begin{tabular}{lccccc}
\toprule
\textbf{Method} & \textbf{Spatial} & \textbf{Object} & \textbf{Goal} & \textbf{Long} & \textbf{Avg.} \\
\midrule
LaRA-VLA ($\theta_\textnormal{ao}$) 
& 96.4\% & \textbf{98.6\%} & \textbf{99.8\%} & \textbf{96.6\%} & \textbf{97.9\%} \\
StarVLA ($\theta_\textnormal{ft}$) 
& \textbf{96.8\%} & 86.6\% & 98.2\% & \underline{96.4\%} & 94.5\% \\
\method~(ours) 
& \underline{96.6\%} & \underline{98.2\%} & \underline{99.2\%} & 94.4\% & \underline{97.1\%} \\
\bottomrule
\end{tabular}
\end{table}

\Cref{tab:libero_laravla,tab:robotwin} shows that \method~is effective for $\theta_\textnormal{pt}$ with both autoregressive architectures (\textit{e.g.,} OpenVLA) and flow-matching architectures (\textit{e.g.,} StarVLA and $\pi_{0.5}$), obtaining consistent improvements of success rates and achieving similar performance as $\theta_\textnormal{ao}$.
Given that the flow-matching expert is typically initialized prior to finetuning, we evaluate two variants: one that merges only the parameters of the Vision-Language Model (VLM), and another that merges both the VLM and the action expert. 
\Cref{tab:robotwin} shows that both variants reach higher success rates over the regularly finetuned $\pi_{0.5}$, and merging both the parameters of VLM and the action expert yields relatively better performance.

\subsection{Determinants of Capability Vector Quality (RQ4)}
\label{sec:determinants}


Given our \method's effectiveness, it is important to further explore how to obtain higher-quality capability vectors in order to achieve a better meta model $\theta_\textnormal{meta}$.
Considering that visual perception is a critical factor in capability transferring and determining whether a VLA model can output accurate actions, we focus on investigating capability vectors obtained from $\mathcal{D}_\textnormal{ext}$ datasets with different visual characteristics.
\begin{table}[t]
\centering
\small
\caption{Visual richness of different $\mathcal{D}_\textnormal{ext}$. BGs \& Obj. Pairs refers to the number of distinct combinations formed by different backgrounds and object pairs.}

\begin{tabular}{cccccc}
\toprule
\textbf{Benchmark} & \multicolumn{3}{c}{\textbf{LIBERO}} & \multicolumn{2}{c}{\textbf{RoboTwin}} \\
\midrule
\textbf{Dataset} & Spatial & Long & 90 & Clean & Randomized \\
\midrule
Tasks & 10 & 10 & 90 & 5 & 5 \\
Backgrounds & 1 & 3 & 3 & 1 & 10k \\
BGs \& Obj. Pairs & 10 & 10 & 90 & 5 & 50k \\
Pairs per Task & 1 & 1 & 1 & 1 & 10k \\
\bottomrule
\end{tabular}
\label{tab:statistics}
\end{table}

\noindent
\textbf{\textit{Finding 5: Diverse task-irrelevant visual features yield high-quality capability vectors.}}
In \Cref{fig:richness_compare}, we mainly compare the $\mathcal{D}_\textnormal{ext}$ with clean backgrounds and randomized backgrounds individually, which represent different levels of \textbf{data diversity} under the same data volume.
Data diversity is defined as the diversity of data instances associated with each task-specific dataset \citep{xing2025shortcut}. 
In our case, data diversity is positively correlated with \textit{Pairs per Task} in \Cref{tab:statistics}. Specifically, when the number of tasks is fixed, richer variations in backgrounds and objects lead to a higher number of \textit{pairs per task}, and consequently, greater data diversity.

As shown in \Cref{fig:richness_compare}, under the same data scale and number of tasks, $\mathcal{D}_\textnormal{ext}$ with randomized backgrounds yields a significantly higher success rate than its clean-background counterpart. 
This indicates that a higher data diversity corresponds to higher-quality capability vectors.
This improvement can be attributed to the fact that higher data diversity fosters more robust and generalizable spatial understanding capabilities. 
Such capabilities are subsequently transferred to OOD domains through the capability vectors.

\noindent
\textbf{\textit{Finding 6: Task-relevant visual cues in $\mathcal{D}_\textnormal{ext}$ can lead to shortcut learning and degrade capability vectors.}}
In the grey part of \Cref{tab:robotwin}, we investigate the model performance on an OOD target dataset $\mathcal{D}_\textnormal{down}$: \{RoboTwin 2.0\}, and $\mathcal{D}_\textnormal{ext}$: \{LIBERO-Spatial, Long, and 90\}.
As summarized in \Cref{tab:statistics}, LIBERO-Spatial and LIBERO-Long each contain 10 tasks, whereas LIBERO-90 comprises 90 tasks. In terms of visual diversity, LIBERO-Spatial includes only a single background, while both LIBERO-Long and LIBERO-90 contain three distinct backgrounds. Notably, all three datasets share the same number of \textit{pairs per task}, indicating a comparable level of data diversity.
We therefore consider an alternative factor, namely \textbf{data disparity}, which is introduced to quantify the heterogeneity across task-specific datasets. Data disparity $S_\textnormal{disparity}$ is defined as the inverse of the expected pairwise similarity between datasets. Under the same \textit{pairs-per-task} setting, increasing variations in backgrounds and objects lead to higher data disparity, yielding the ordering: $S_\textnormal{disparity}^\textnormal{spatial} < S_\textnormal{disparity}^\textnormal{Long} < S_\textnormal{disparity}^\textnormal{90}.$
This trend is inversely correlated with the observed success rates (SR): $\mathrm{SR}_{\text{Spatial}} > \mathrm{SR}_{\text{Long}} > \mathrm{SR}_{90}.$

We hypothesize that task-specific backgrounds and objects may induce spurious correlations, such that the finetuning process of Spatial Forcing implicitly prioritizes the learning of simpler, higher-variance patterns \citep{arpit2017closer}. 
Higher data disparity implies greater variance across tasks; when the disparity introduced by backgrounds and objects dominates task-relevant characteristics (e.g., object positions or dynamics), the model tends to preferentially learn these high-variance features, resulting in \textit{shortcut learning}. 
Therefore, avoiding shortcut learning is crucial for inducing a meaningful capability vector.

\subsection{Real-world Study (RQ5)}
\label{sec:real}


\begin{figure*}[t]
    \centering
    \includegraphics[width=0.9\linewidth]{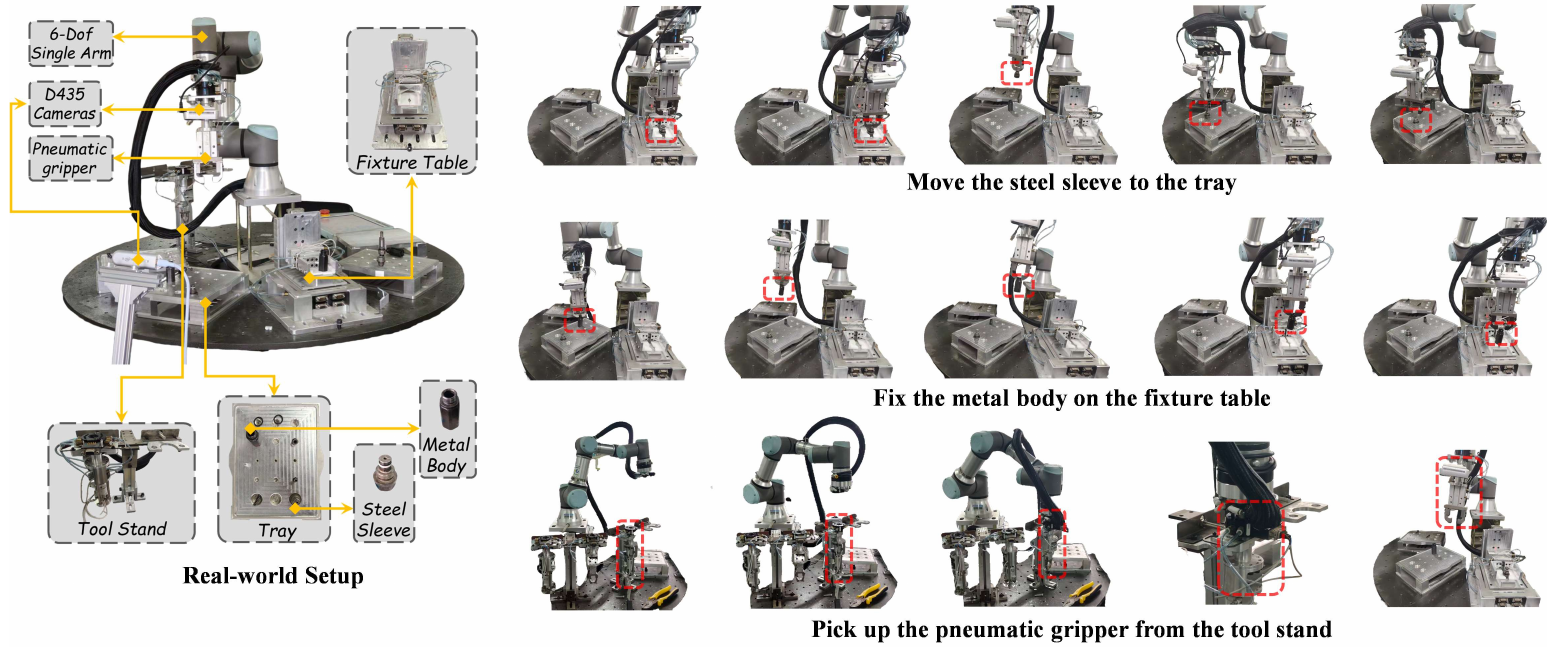}
    \vspace{-0.3cm}
    \caption{
    \textbf{Real-world setup on industrial tasks on UR3 robot.} 
    }
    \label{fig:real_task_ur3_main}
    \vspace{-0.3cm}
\end{figure*}

\begin{figure*}[t]
    \centering
    \includegraphics[width=0.86\linewidth]{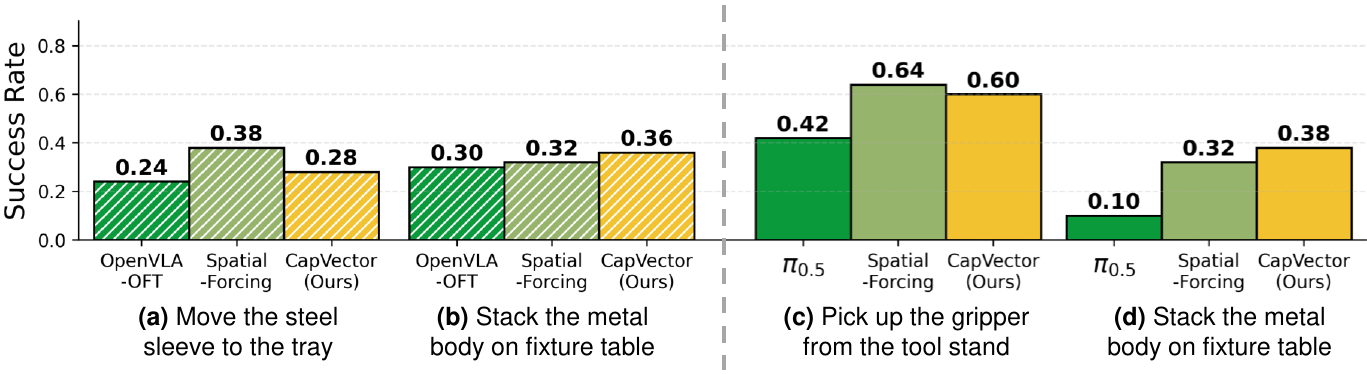}
    \vspace{-0.3cm}
    \caption{
    \textbf{Real-world experiments on industrial tasks on UR3 robots.} 
    $\theta_\textnormal{ft}$: \{$\pi_{0.5}$ and OpenVLA-OFT\}. $\theta_\textnormal{ao}$: \{Spatial Forcing\}.
    $\mathcal{D}_\textnormal{ext}$: \{LIBERO-Spatial\}.
    $\mathcal{D}_\textnormal{down}$: \{3 real-world industrial tasks\}.
    }
    \label{fig:real_task}
\end{figure*}

\textbf{Settings.}
We verify the effectiveness of our \method~on the real-world hardware platform. \Cref{fig:real_task_ur3_main} shows that the platform serves as a flexible assembly testbed designed for industrial applications. 
We design a comprehensive suite of tasks that encompasses common robotic manipulation scenarios in industrial production. 
We collect 100 episodes per task and finetune on all tasks together.
During evaluation, models are tested over 100 rollouts per task to obtain reliable success rates.

\textbf{\textit{Finding 7:}} \method~\textbf{\textit{can realize sim-to-real transferring and is robust and general for diverse embodiments and scenes.}}
To verify whether the extra spatial capabilities extracted from simulation can directly generalize to the physical world, we utilized the capability vectors merged solely from LIBERO-Spatial and applied them to our real-world training. The quantitative results shown in \Cref{fig:real_task} demonstrate that \method~yields substantial improvements over the standard baselines across all tasks, notably even surpassing Spatial Forcing on some tasks.

These results suggest that the spatial perception capabilities inherent in the capability vectors are environment-agnostic to a certain degree. These vectors capture fundamental geometric cues rather than overfitting to the domain-specific visual textures. The finding highlights the practical value of \method, as it enables the utilization of scalable simulation datasets to extract the transferable capabilities, while avoiding the complexity of applying extra auxiliary objectives in the real world.

\begin{figure}[t]
    \centering
    \includegraphics[width=\linewidth]{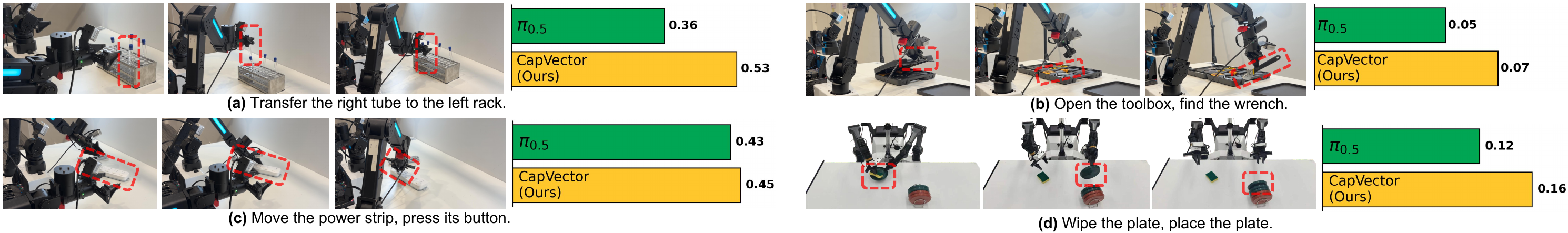}
    \caption{
    \textbf{Cross-embodiment deployment of capability vectors on ARX Lift 2 and AgileX Cobot out of the box.} 
    $\theta_\textnormal{ft}$: \{$\pi_{0.5}$\}. $\theta_\textnormal{ao}$: \{Spatial Forcing\}.
    $\mathcal{D}_\textnormal{ext}$: \{LIBERO-Spatial\}.
    $\mathcal{D}_\textnormal{down}$: \{4 external real-world tasks\}.
    }
    \label{fig:real_task_external}
\end{figure}

To evaluate system robustness and cross-embodiment performance, we share the same capability vectors' weight with 2 external collaborators, allow them to fine-tune the $\theta_\textnormal{meta}$ on 2 embodiments (ARX Lift 2 and AgileX Cobot) and 4 complex tasks, and get performance numbers from them.
\Cref{fig:real_task_external} shows that our method consistently outperforms the base model to varying degrees. 
In particular, \Cref{fig:real_task_external}~(a) requires the model to sequentially transfer four test tubes, posing a significant challenge to its long-horizon precise manipulation capability. For this task, our method improves the success rate from 0.36 to 0.53.
This demonstrates that our capability vectors have superior out-of-the-box performance and our \method~can be a general strategy to improve pretrained models and finetuning processes. Detailed setup of these 2 embodiments is shown in \cref{appendix:real_world}.



\section{Related Work}
\label{sec:relatedwork}
\paragraph{SFT Strategies for VLAs.}
Recent works have increasingly focused on advancing generalist VLAs \citep{RT-2, kim2025openvla, black2024pi_0, black2025pi, wen2025diffusionvla, bjorck2025gr00t, bai2026hex, song2025pd, bai2025towards} trained on large-scale robot data \citep{o2024open,khazatsky2024droid, wu2025robocoin} based on VLMs \citep{beyer2024paligemma, yang2025qwen3}. 
However, standard SFT on these foundation models is often insufficient for the model to quickly achieve high performance on new tasks. 
Therefore, many auxiliary-objective SFT strategies are proposed. OpenVLA-OFT \citep{kim2025fine} improves performance and inference efficiency via optimized action decoding strategies and learning objectives, while FLARE \citep{flare} and FRAPPE \citep {zhao2026frappe} introduces future latent representation alignment for implicit world modeling. Spatial Forcing \citep {li2025spatial} enhances spatial abilities by imposing spatial alignment constraints. LaRA-VLA \citep{laravla}, $\textnormal{LaST}_0$ \citep{liu2026last}, and DualCoT-VLA \citep{zhong2026dualcot} internalize multi-modal CoT into continuous latents for efficient reasoning.
Although finetuning with auxiliary objectives provides enhanced performance and efficiency, they often introduce additional forward passes.
In this paper, we avoid the overhead through capability extraction and model merging.

\paragraph{Model Merging.}
Model merging combines parameters of two distinct models to reuse knowledge and improve robustness. Prior work in LLMs and VLMs \citep{wang2024localizing, yadav2023ties, yadav2024matters, nasery2025pleas, lu2025fine, jang2024model} shows that simple weight interpolation or averaging can effectively combine task-specific skills and mitigate distribution shifts. O-LoRA \citep{o-lora} further mitigates catastrophic forgetting by orthogonal low-rank updates.
However, extending model merging to VLA models has received limited attention. ReVLA \citep{dey2025revla} alleviates visual catastrophic forgetting via vision backbone reversal. RETAIN \citep{yadav2025robust} enables robust adaptation in low-data regimes by interpolating pretrained and finetuned VLA checkpoints. MergeVLA \citep{fu2025mergevla} further proposes a cross-skill composition strategy.
Unlike simple model ensembling, we extract capability vectors via arithmetic operations on model parameters, followed by model merging to achieve capability transfer across models.

\section{Conclusion}
\label{sec:conclusion}
We introduce the concept of capability vectors to represent the gains in general capabilities acquired during finetuning, and propose a pipeline that integrates these vectors. By merging capability vectors with pretrained models, we construct a capability-enhanced meta model. \method~achieves the simplicity and computational efficiency of standard SFT with the high performance of auxiliary-objective SFT. 
Through a series of experiments, we systematically analyze the effectiveness, versatility, and mechanism of capability vectors. Finally, real-world experiments validate the practical applicability and generalization. 
Overall, this work provides a novel strategy for extracting and transferring gains from finetuning into pretrained models, offering a feasible solution for more efficient and broadly applicable VLA training.

\section{Acknowledgments}
We acknowledge Jiahao Chen and Tongshuo Xu from Tsinghua University for their assistance with the collection of real-world training data.
We acknowledge Shuanghao Bai from Xi'an Jiao Tong University for his assistance with the deployment of baselines.

\bibliographystyle{assets/plainnat}
\bibliography{paper}

\clearpage
\newpage
\onecolumn
\beginappendix
\renewcommand{\thefigure}{S\arabic{figure}}
\renewcommand{\thetable}{S\arabic{table}}
\setcounter{figure}{0}
\setcounter{table}{0}

\section{More Ablations.}
\label{appendix:ablation}

$\alpha$ is utilized to control the merging weight of the capability vectors during the model merging phase, as defined in \cref{eq:cap_merge}. This weight dictates the degree to which the extracted general capabilities $\gamma_{ao}$ are integrated into the pretrained backbone $\theta_{pt}$ to construct the capability-enhanced meta-model $\theta_{meta}$. As shown in \cref{tab: cap_merge_weight}, the model achieves its optimal performance when $\alpha$~=~1.1, which is chosen as the default setting in all other experiments.

\begin{table}[h]
\caption{\textbf{Ablation study for the merging weight $\alpha$ of the capability vectors.} We report the Success rates (\%). $\mathcal{D}_\textnormal{ext}$: \{LIBERO-Spatial\}. $\mathcal{D}_\textnormal{down}$: \{LIBERO-Long\}.}
\centering
\label{tab: cap_merge_weight}
\resizebox{0.48\textwidth}{!}{
\begin{tabular}{ccccccc}
\toprule
\multirow{2}{*}{\makecell[c]{VLA \\ Backbone}} & \multicolumn{6}{c}{Weight $\alpha$} \\
\cmidrule(lr){2-7}
 & 0.5 & 0.7 & 0.9 & 1.1 & 1.3 & 1.5 \\
\midrule
 OpenVLA-OFT & 91.2 & 91.8 & 92.8 & \textbf{94.8} & 90.6 & 92.4 \\
\bottomrule
\end{tabular}
}
\end{table}

\section{Extra Overhead Induced by Orthogonal Loss}
\label{appendix:orth_loss}

\begin{table}[h]
\caption{\textbf{Extra overhead of orthogonal loss.} We report the training computational cost (FLOPs) and GPU memory based on the LoRA tuning of OpenVLA-OFT.}
\centering
\label{tab:reg_loss_cost}
\begin{tabular}{cll}
\toprule
\textbf{Orthogonal Loss} & \textbf{FLOPs} & \textbf{GPU Memory} \\ \midrule
\usym{2718} & 17.9T & 62.8G \\
\usym{2714}  & 17.9T + 0.3G & 62.8G + 0.5G \\ \bottomrule
\end{tabular}
\end{table}

To maintain the capability of the capability vectors during downstream finetuning, we incorporate the orthogonal loss, as illustrated in \Cref{eq:orth_loss}. We evaluate its computational overhead by measuring the computational FLOPs and GPU memory usage, as summarized in \Cref{tab:reg_loss_cost}.
The orthogonal loss introduces a negligible computational overhead, increasing total training FLOPs by merely \textbf{0.3G (<0.002\%)} and GPU memory usage by approximately \textbf{0.5G (<0.8\%)}. Consequently, \method~achieves the superior performance improvement through our orthogonal loss while maintaining simplicity and low resource consumption.

\section{Extra Overhead Comparison between Auxiliary-objective SFT Methods and Ours}
\label{appendix:sf_overhead}

\begin{table}[ht]
\caption{\textbf{Extra overhead of Auxiliary-objective SFT.} We report the training computational cost (FLOPs) and GPU memory based on the LoRA tuning of OpenVLA-OFT. We take Spatial Forcing as the example of auxiliary-objective SFT methods.}
\centering
\label{tab:sf_loss_cost}
\begin{tabular}{lll}
\toprule
\textbf{Method} & \textbf{FLOPs} & \textbf{GPU Memory} \\ \midrule
OpenVLA-OFT & 17.9T & 62.8G \\
+ Spatial Forcing & 17.9T + 5.0T & 62.8G + 10.9G \\
+ \method~(Ours)  & 17.9T + 0.3G & 62.8G + 0.5G \\ \bottomrule
\end{tabular}
\end{table}

As summarized in \Cref{tab:sf_loss_cost}, the auxiliary-objective SFT method (\textit{e.g.}, Spatial Forcing) incurs substantial computational costs because it introduces additional forward passes and auxiliary modules to align the targets. Specifically, adding Spatial Forcing to the OpenVLA-OFT baseline increases training FLOPs by \textbf{5.0T (28\%)} and GPU memory usage by \textbf{10.9G (17\%)}. By contrast, our \method~introduces a negligible overhead while achieving performance comparable to, or even exceeding, full auxiliary-objective finetuned methods across diverse tasks.

\section{Real-world Setup.}
\label{appendix:real_world}

\noindent
\textbf{Data Collection.}
During the data collection phase, we first manually designate a diverse set of intermediate waypoints for each task. 
Subsequently, the robot autonomously generated a wide variety of trajectories by randomly selecting from these waypoints. 
Finally, we collected a total of 100 episodes per task. 
\begin{figure*}[t]
    \centering
    \includegraphics[width=0.7\linewidth]{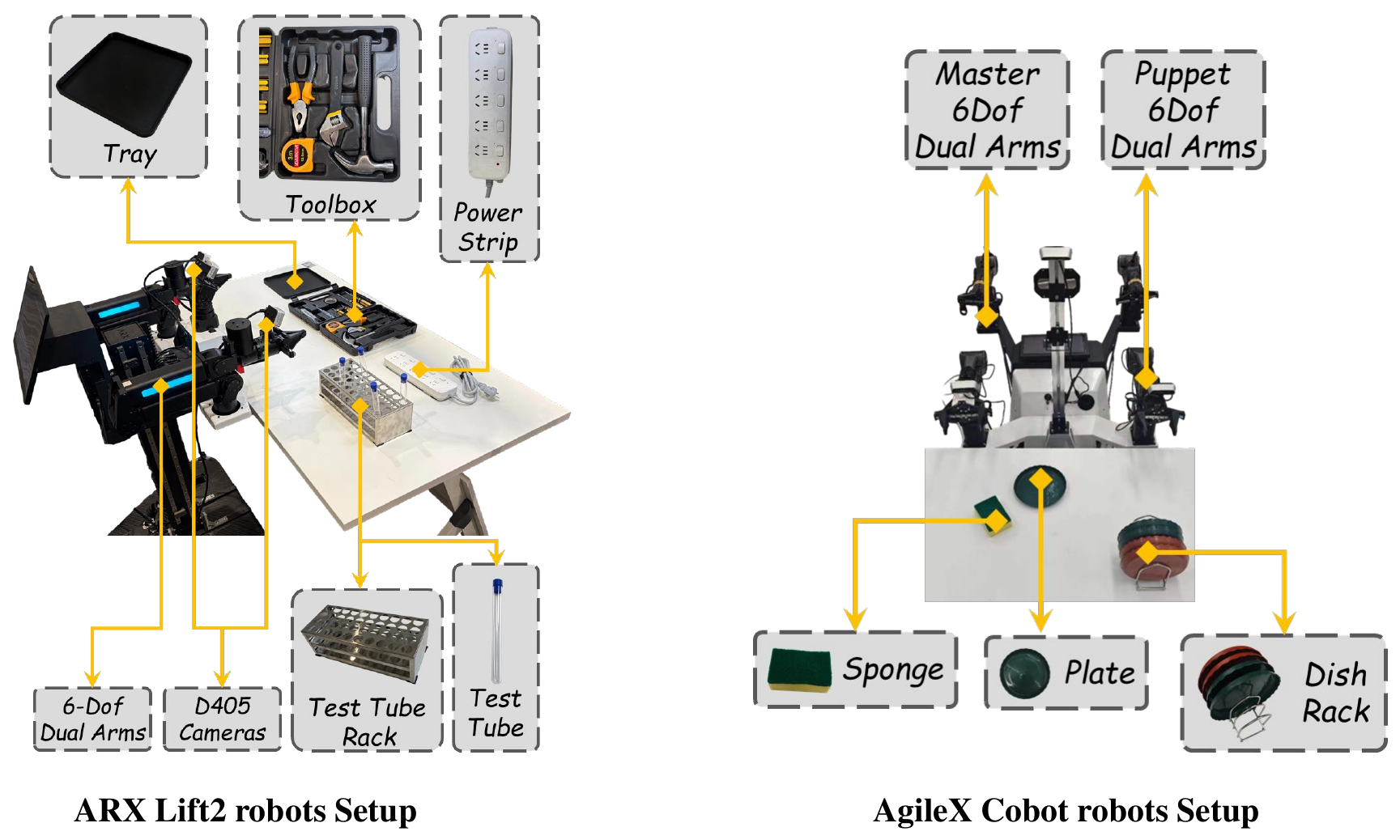}
    \vspace{-0.3cm}
    \caption{
    \textbf{Real-world setup of the cross-embodiment deployment tasks on ARX Lift2 and Agilex Cobot robots.} 
    }
    \label{fig:real_task_lift2}
\end{figure*}

\textbf{Robot and Task Setup.}
In \cref{sec:real}, we deploy our models across three distinct hardware platforms to evaluate both specific real-world performance and cross-embodiment generalization. Here we show those two external collaborators' real-world robot and task setups in \cref{fig:real_task_lift2}.
\begin{itemize}
    \item ARX Lift 2: This platform utilizes a 6-DoF dual-arm system with D405 cameras. The task workspace is configured with objects like a test tube rack, test tubes, a toolbox, a tray, and a power strip to evaluate complex, multi-stage manipulations.
    \item AgileX Cobot: This setup consists of a teleoperation-ready system featuring master and puppet 6-DoF dual arms. The workspace includes a dish rack, plate, and sponge, designed to evaluate everyday dexterous tasks.
\end{itemize}

\section{Limitations.}
\label{appendix:limitations}
Overall, post-training for robot foundation models can be broadly categorized into two paradigms: supervised fine-tuning (SFT) and reinforcement learning (RL). In this work, we primarily focus on capability vectors obtained through SFT, while leaving the acquisition of capability vectors in RL settings to future work.

\section{Baselines.}
\label{appendix:baseline}

\textbf{OpenVLA-OFT \citep{kim2025fine}.} OpenVLA-OFT proposes an Optimized finetuning (OFT) strategy aimed at improving both the performance and inference efficiency of vision-language-action (VLA) models when adapted to specific robotic tasks. This approach combines parallel decoding, action chunking, continuous action representations, and a straightforward L1 regression loss to accelerate inference. For tasks demanding fine-grained language comprehension, OFT is enhanced with FiLM (Feature-wise Linear Modulation) \citep{perez2018film} to reinforce language grounding. On the LIBERO simulation benchmark \citep{liu2023libero}, OpenVLA-OFT achieves a 97.1\% success rate while increasing action generation speed by 26 times. In real-world experiments with a bimanual ALOHA robot \citep{zhao2024aloha}, it surpasses strong finetuned VLAs such as $\pi_0$ \citep{black2024pi_0} and RDT-1B \citep{rdt}, as well as policies trained from scratch, achieving up to a 15\% absolute improvement in average success rate on dexterous manipulation tasks.

\textbf{$\pi_{0.5}$ \citep{black2025pi}.} $\pi_{0.5}$ is a vision-language-action (VLA) model designed to achieve open-world generalization for real-world robotic manipulation. Building on $\pi_{0}$, $\pi_{0.5}$ adopts a co-training framework that leverages highly heterogeneous data sources, including demonstrations from multiple robot platforms, high-level semantic prediction tasks, web-scale multimodal data, and language supervision. The model follows a hierarchical architecture that first infers high-level semantic subtasks and subsequently predicts low-level action chunks, enabling long-horizon and multi-stage manipulation. Despite relying predominantly on non-mobile-manipulation data during training, $\pi_{0.5}$ demonstrates strong generalization to unseen homes and complex real-world tasks, such as household cleaning and dexterous object rearrangement.

\textbf{StarVLA \citep{starvla}.} StarVLA introduces a modular, Lego-like unified framework to address the severe fragmentation across architectures, codebases, and evaluation protocols in VLA research. It implements a shared "backbone-action-head" abstraction that allows researchers to seamlessly interchange diverse vision-language or world-model backbones (\textit{e.g.}, Qwen-VL \citep{yang2025qwen3} and Cosmos \citep{cosmos-policy}) with four representative action-decoding paradigms (\textit{e.g.}, autoregressive tokenization, flow-matching). This unified architecture is further strengthened by reusable, paradigm-agnostic training strategies for multimodal co-training and a standardized server-client interface for cross-benchmark evaluation. Across major benchmarks including LIBERO, SimplerEnv, and RoboTwin 2.0, StarVLA provides fully reproducible single-benchmark training recipes that match or surpass prior state-of-the-art methods, significantly lowering the barrier for principled comparison and practical real-robot deployment.


\textbf{Spatial Forcing \citep{li2025spatial}.} Spatial Forcing (SF) introduces a straightforward yet effective alignment strategy to enhance spatial reasoning in VLA models. SF implicitly guides VLAs to acquire 3D spatial comprehension by aligning intermediate visual embeddings with geometric representations extracted from pretrained 3D foundation models. This alignment improves action precision without requiring explicit 3D sensor inputs or depth estimators. On the LIBERO \citep{liu2023libero} and RoboTwin \citep{robotwin} benchmarks, SF surpasses strong 2D- and 3D-based VLA baselines, achieving state-of-the-art performance while accelerating training by up to 3.8× and demonstrating improved data efficiency across diverse robotic tasks.

\textbf{LaRA-VLA \citep{laravla}.} Latent Reasoning VLA (LaRA-VLA) introduces a unified framework that internalizes multi-modal chain-of-thought (CoT) reasoning into continuous latent representations for embodied action. LaRA-VLA implicitly guides the model to reason in latent space via a curriculum-based training paradigm, progressively transitioning from explicit textual and visual CoT supervision to pure latent reasoning. This alignment resolves the representational mismatch between discrete reasoning tokens and continuous control, effectively eliminating explicit CoT generation overhead during inference. On LIBERO \citep{liu2023libero} simulation benchmarks and long-horizon real-robot manipulation tasks, LaRA-VLA consistently outperforms state-of-the-art VLA baselines, achieving superior performance while reducing inference latency by up to 90\% for efficient real-time control.



\end{document}